\documentclass[tinyml]{acmart}

\AtBeginDocument{%
  \providecommand\BibTeX{{%
    \normalfont B\kern-0.5em{\scshape i\kern-0.25em b}\kern-0.8em\TeX}}}

\setcopyright{rightsretained}
\copyrightyear{2022}
\acmYear{2022}

\usepackage{multirow}
\usepackage{graphicx}
\usepackage{enumitem}
\usepackage{cleveref}
\usepackage{subfig}

\newcommand{\sys}{TinyM$^2$Net}
\begin{document}

\title{TinyM$^2$Net: A Flexible System Algorithm Co-designed \underline{M}ulti\underline{m}odal Learning Framework for Tiny Devices}

\author{Hasib-Al Rashid}
\authornote{Both authors contributed equally to this research.}
\email{hrashid1@umbc.edu}
\affiliation{%
  \institution{University of Maryland, Baltimore County}
  \city{Baltimore}
  \state{Maryland}
  \country{USA}
  \postcode{21250}
}

\author{Pretom Roy Ovi}
\authornotemark[1]
\email{povi1@umbc.edu}
\affiliation{
  \institution{University of Maryland, Baltimore County}
  \streetaddress{1000 Hilltop Circle}
  \city{Baltimore}
  \state{Maryland}
  \country{USA}
  \postcode{21250}
}

\author{Carl Busart}
\affiliation{%
  \institution{U.S. Army Research Laboratory}
  \city{Adelphi}
  \state{Maryland}
  \country{USA}
}

\author{Aryya Gangopadhyay}
\affiliation{
  \institution{University of Maryland, Baltimore County}
  \streetaddress{1000 Hilltop Circle}
  \city{Baltimore}
  \state{Maryland}
  \country{USA}
  \postcode{21250}
}

\author{Tinoosh Mohsenin}
\affiliation{
  \institution{University of Maryland, Baltimore County}
  \streetaddress{1000 Hilltop Circle}
  \city{Baltimore}
  \state{Maryland}
  \country{USA}
  \postcode{21250}
}

\renewcommand{\shortauthors}{Rashid and Ovi, et al.}
\begin{abstract}
With the emergence of Artificial Intelligence (AI), new attention has been given to implement AI algorithms on resource constrained tiny devices to expand the application domain of IoT. Multimodal Learning has recently become very popular with the classification task due to its impressive performance for both image and audio event classification. This paper presents \emph{\sys{}} - a flexible system algorithm co-designed multimodal learning framework for resource constrained tiny devices. The framework was designed to be evaluated on two different case-studies: COVID-19 detection from multimodal audio recordings and battle field object detection from multimodal images and audios. In order to compress the model to implement on tiny devices, substantial network architecture optimization and mixed precision quantization were performed (mixed 8-bit and 4-bit). \emph{\sys{}} shows that even a tiny multimodal learning model can improve the classification performance than that of any unimodal frameworks. The most compressed \emph{\sys{}} achieves 88.4\% COVID-19 detection accuracy (14.5\% improvement from unimodal base model) and 96.8\% battle field object detection accuracy (3.9\% improvement from unimodal base model). Finally, we test our \emph{\sys{}} models on a Raspberry Pi 4 to see how they perform when deployed to a resource constrained tiny device.
\end{abstract}



\maketitle

\section{Introduction}
Artificial Intelligence (AI) has a huge impact on our daily lives now-a-days. In our daily lives, AI has brought convenience and ease of use to the table. The AI devices are now able to perform computationally intensive tasks and eliminate human error from the system to a large extent, making this convenience possible. We see AI techniques and devices being used in domains such as medical diagnosis, security and combat fields, robotics, vision analytics, knowledge reasoning, navigation, etc. today. To integrate AI in our day-to-day life, it is being implemented on resource constrained mobile and edge platforms. With the exponential growth of resource constrained micro-controller (MCU) and micro-processor (MPU) powered devices, a new generation of neural networks has emerged, one that is smaller in size and more concerned with model efficiency than model accuracy. These low-cost, low-energy MCUs and MPUs open up a whole new world of tiny machine learning (TinyML) possibilities. We can directly do data analytics near the sensor by running deep learning models on very tiny devices, greatly expanding the field of AI applications.

Modern IoT and wearable devices, such as activity trackers, environmental sensors, images, and audio sensors can generate large volumes of data on a regular basis. Modern AI is increasingly reliant on data from numerous sources in order to produce more accurate findings. Learning process for human is multimodal. We human can take our decision by processing different modalities of data. To mimic human-like behavior, AI algorithms should integrate multimodal data as well. Multimodal learning combines disparate, heterogeneous data from a variety of sensors and data sources into a single model. In contrast to standard unimodal learning systems, multimodal systems can convey complimentary information about one another, which becomes apparent only when both are integrated into the learning process. Thus, learning-based systems that incorporate data from many modalities can generate more robust inference or even novel insights, which would be unachievable in a unimodal system. Multimodal learning has two key advantages. To begin, several sensors observing the same data can produce more robust predictions, as recognizing changes in it may need the presence of both modalities. Second, the integration of many sensors enables the capture of complementing data or trends that individual modalities may miss. However, increased model parameters and computations limit multimodal learning to be adopted for resource constrainded edge and tiny ML applications.

Commodity MCUs and MPUs have a very limited resource in terms of memory (SRAM) and storage (Flash) budget. A typical MCU has an SRAM of less than 512kB, which is insufficient for installing the majority of off-the-shelf deep learning networks. Even on more capable hardware such as the Raspberry Pi 4, configuring inference to run in the L2 cache (1MB) can dramatically increase energy efficiency. These new issues add to the difficulty of performing efficient multimodal learning inference with a low peak memory consumption. In this paper, we address this challenge and implement multimodal learning on tiny hardware. We take advantages of state-of-the-art compression techniques and combined them with computationally relaxed layers to implement energy efficient multimodal learning on tiny processing hardware. We proposed a flexible system algorithm co-designed framework \emph \sys{} which is re-configurable in terms of input data modality and data shapes, number of layers, filter sizes etc. hyper-parameters for the sake of application requirements. We evaluated \emph \sys{} with two different case-studies: audio processing with multimodal audios and object detection with multimodal images and audios. \emph \sys{} is then implemented on commodity tiny MPU, Raspberry Pi 4 to measure real-time performance on tiny hardware. The main contributions of this paper are as follows:

\begin{itemize}[leftmargin=0.25in]
 \item Propose \emph \sys{}, a novel flexible system-algorithm co-designed multimodal learning framework for resource constrained devices. \emph{\sys{}} that can take multimodal inputs (images and audios) and be re-configured for application specific requirements. \emph{\sys{}} allows the system and algorithms to quickly integrate new sensors data that are customized to various types of scenarios.
  \item Perform network architecture optimization, and mixed-precision quantization with the purpose of decreasing computation complexity and memory size for resource constrained hardware implementation while maintaining accuracy.
 \item Evaluated proposed \emph \sys{} for two different case-studies. \emph{Case-study 1} includes Covid-19 detection from multimodal cough and speech audio recordings. \emph{Case-study 2} includes battlefield object detection using multimodal images and audios.
 \item Implement \emph \sys{} on commodity microprocessor unit, Raspberry Pi 4. We measured inference time while it was in use, as well as providing the appropriate power profiling to ensure that our system is adaptable. To be called a real-time implementable tinyml system, \emph \sys{} meets all of the requirements.

\end{itemize}

\section{Related Works}
Authors in \cite{mazumder2021survey} presented a high level overview on the optimization techniques of deep neural networks (DNNs) for tinyML on device inferences. TinyML model optimization includes different algorithms of Parameter Search, Sparsification and Quantization techniques. Element-wise pruning\cite{meng2021fixyfpga}, Structured Pruning \cite{anwar2017structured,lemaire2019structured} these techniques reduces the unimportant weights and compress the models to be implemented on tinyML devices. Extreme low precision quantization \cite{liang2018fp, alemdar2017ternary} and mixed precision quantization \cite{yao2021hawq, hubara2020improving, wang2019haq, wu2018mixed, hosseini2021qs} is adopted by the researchers to decrease the memory requirements of the DNN models. MCUNet \cite{lin2020mcunet}, MicroNets \cite{banbury2021micronets} are proposed to deploy DNN models on micro-controller units (MCU). 
Recently, multimodal learning attracts researchers to improve the classification accuracy of the models integrating different modalities of data fusion \cite{bouchey2021multimodal,abdullah2021multimodal,ma2021smil, 9556252}. However, implementation of multimodal learning into resource constrained tiny hardware is very limited due to its large model sizes. We present a novel multimodal learning framework \emph{\sys{}} which is system algorithm co-designed and different model compression techniques were implemented to compress the large multimodal models to be implemented on tiny devices.

\section{\protect\sys{} Framework}

\begin{figure*}[htb]
\centering
\includegraphics[width=0.8\textwidth]{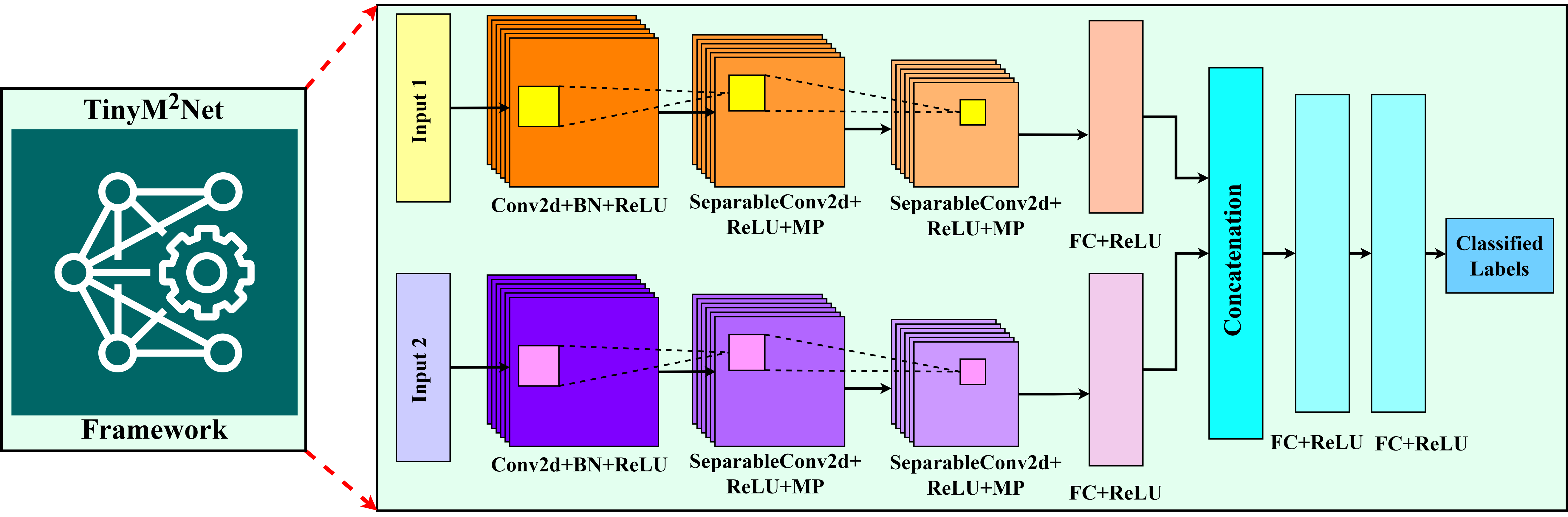}
\caption{\small The proposed \emph{\sys{}} framework for multimodal learning on tiny hardwares. \emph{\sys{}} is flexible in terms of number of input, number of layers and hyper-parameters based on application specific requirement. New data modality suited for various settings can be readily included into the device. Some of the input information is images, other input data can be auditory.}
\label{framework}
\end{figure*}

Figure \ref{framework} shows the proposed \emph{\sys{}} framework along with its detailed architecture. Based on the case studies we mention in section \ref{eve}, \emph{\sys{}} is able to integrate two different modalities of data and classify them. Proposed \emph{\sys{}} is designed based on mainly convolutional neural networks (CNN). CNN performed very promising with images and audio data classification previously which is the reason behind choosing CNN as our base model.    

We will evaluate \emph{\sys{}} on two different case studies described in details in section \ref{eve}. Case-study 1 is detecting COVID-19 signatures using two different modalities of audio recordings, cough sound and speech sound. Case-study 2 is based on detecting battle field object  using images and audios. While processing audio data, we have divided the whole audio into shorted window frames, the size of those frames are variable based on application requirements. Then the window frames are converted into Mel-Frequency Cepstral Coefficients (MFCCs) spectrograms. In the next step, two different data modalities, whether be images or audios, are sent to the CNN layers for feature extraction. The number of layers of our CNN layers can be adjusted to suit application specific requirements. Maxpooling layers is used to reduce the size of the feature map. A number of fully interconnected layers are applied once the output is flattened to achieve the needed tiny feature map size in order to isolate enough information with linkages between nodes. The outputs from two parallel feature extraction layers from two data modalities are then concatenated and  processed through fully connected layers to produce the final label. The activation function for each layer is a rectified linear unit (ReLU). Softmax activation function is used to generate a probability distribution for the final layer.

\section{Model Compression for Tiny Devices}
Traditional CNN models are very notorious for being bulky in terms of memory and computation requirements. To implement CNN on low powered embedded tiny devices, researchers proposed various compression techniques which results in highly optimized CNN models. Our proposed \emph{\sys{}} adopts different model compression techniques which optimizes the network architectures and memory requirements. \emph{\sys{}} adopts Depthwise Separable CNN (DS-CNN) to reduce the computation from traditional CNN layers. To have improvement on memory requirements, \emph{\sys{}} adopts low precision and mixed-precision (MP) quantization. We emphasize on MP quantization as uniform low precision quantization degrades model accuracy.

\subsection{Network Architecture Optimization with DS-CNN}
Figure \ref{conv-dsconv} presents the conventions and the techniques that is done in traditional CNN and DS-CNN . In traditional CNN, if the input is of size ${D_f \times D_f \times M}$ and $N$ is the number of filters having a size ${D_k \times D_k \times M}$ then output of this layer without zero padding applied is of size $D_p \times D_p \times M$. If the stride for the convolution is $S$ then $D_p$ is determined by the following equation:
\begin{equation}\label{convout}
D_p = \frac {D_f - D_k }{S} + 1
\end{equation}
In this layer, the filter convolves over the input by performing element wise multiplication and summing all the values. A very important note is that depth of the filter is always same as depth of the input given to this layer. The computational cost for traditional convolution layer is $M \times D_k^2 \times D_p^2 \times N$ \cite{howard2017mobilenets}.

\begin{figure*}[htb]
\centering
\includegraphics[width=0.8\textwidth]{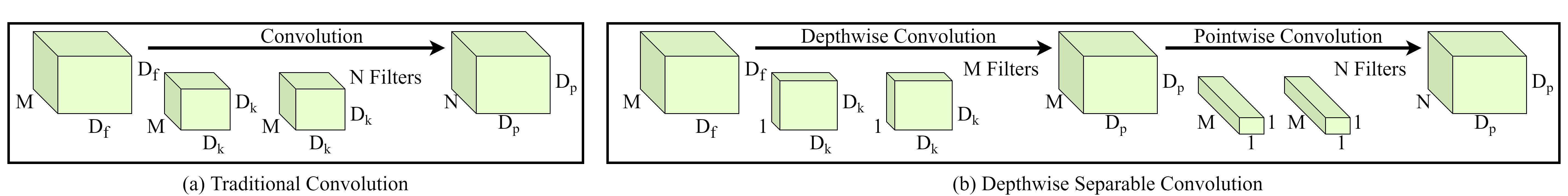}
\caption{\small Detailed operations inside traditional CNN and DS-CNN.}
\label{conv-dsconv}
\end{figure*}

\begin{table}[]
\centering
\caption{\small Number of parameter and required computations equations for different types of convolution layers.}
\resizebox{0.49\textwidth}{!}{%
\begin{tabular}{|c|c|c|c|c|c|c|}
\hline
CNN types & No. of Parameters & No. of Computations \\ \hline
Traditional & $M \times D_k^2 \times N$ & $M \times D_k^2 \times D_p^2 \times N$ \\ \hline
Depthwise Separable & $M \times D_k^2 + M \times N $ & $M \times D_p^2 \times D_k^2 + M \times D_p^2 \times N $ \\ \hline
\end{tabular}
\vspace{-2ex}}
\label{parameters}
\end{table}

\par Depthwise Separable convolution is a combination of depthwise and pointwise convolution \cite{kaiser2017depthwise}. In contrast to traditional CNNs, which apply convolution to all $M$ channels at once, depthwise operations only apply convolution to a single channel at a time. So here the filters/kernels will be of size $D_k \times D_k \times 1$. As there are $M$ channels at the input, $M$ numbers of such filters are needed. This will produce a output of size $D_p \times D_p \times M$. A single convolution operation require $D_k \times D_k$ multiplications. Since the filter are slided by $D_p \times D_p$ times across all the $M$ channels. The total number of computation for one depthwise convolution comes to be $M \times D_p^2 \times D_k^2$. In point-wise operation, a $1\times1$ convolution is applied on the $M$ channels. The filter shape for this operation will be $1 \times 1 \times M$. If we use $N$ such filters, the output shape becomes $D_p \times D_p \times N$. One convolution operation in this needs $1 \times M$ multiplications. The total number of operations for one pointwise convolution is $M \times D_p^2 \times N$. Therefore, total computational cost of one depthwise separable convolution is $M \times D_p^2 \times D_k^2 + M \times D_p^2 \times N $ \cite{howard2017mobilenets}. Table \ref{parameters} shows the equations required for calculating the number of parameters and number of computations for each of the traditional CNN and DS-CNN layers. Here $D_k \times D_k$ is the size of the filter, $D_p \times D_p$ is the size of the output, $M$ is number the of input channels and $N$ is the number of output channels.

\subsection{Model Quantization}
To have lesser memory requirements, model quantization is now attracting to the researchers to design tinyML models. The accuracy of a model can be significantly degraded if it is uniformly quantized to low bit precision. It is possible to address this with mixed-precision quantization in which each layer is quantized with different bit precision. The main idea behind mixed precision quantization is to keep sensitive layers at higher precision and insensitive layers at lower precision. However, the search space for this bit setting grows exponentially as the number of layers increase, which is challenging. Various methods have been offered to deal with this enormous search area. Reinforcement learning (RL) and Neural Architecture Search (NAS) have recently been presented as efficient approaches for searching the search space. As a result, these methods \cite{wang2019haq, wu2018mixed, hosseini2021qs} often require a huge amount of computational resources and their performance is highly dependent on hyperparameters and even initialization. An algorithm known as Integer Linear Programming (ILP) is employed in \cite{yao2021hawq, hubara2020improving}. ILP is very light weight and gives result within second. We adopted ILP and formulated our problem following the methodology described in \cite{yao2021hawq}, simplifying some constraints to get the mixed precision settings for our \emph{\sys{}}. ILP equations were solved using a python module called Pulp.

To tackle the accuracy degradation with extreme low bit precision quantization, we chose two different bit precision settings ($X = 2 $), INT4 and INT8 for our \emph{\sys{}} framework. As our \emph{\sys{}} is flexible in terms of number of layers, for a model with $Y$ layers, the search space for ILP becomes $X^Y$. ILP will find the best bit precision choices from $X^Y$ search spaces to have optimal trade-off between model perturbation $\Omega$ and user specific constraints i.e. model size and Bit Operations, BOPS. Each of these bit-precision options has the potential to produce different model perturbation. We assumed the perturbation for each layer are independent to each other \cite{yao2021hawq} (i.e., $\Omega$ = $\sum_{i=1}^{Y} \Omega_{i}^{x_i}$, where $\Omega_{i}^{x_i}$ is the perturbation of i-th layer with $x_i$ bit). This enables us to pre-calculate the sensitivity of each layer independently, with only $XY$ computations required. Hessian based perturbation, presented in \cite{yao2021hawq} is used as sensitivity metric. Minimizing this sensitivity, ILP tries to find the right bit precision settings. The ILP equations would be as follows:
Objective:
\begin{equation}
\label{obj}
\small
\min{\left\{x_{i}\right\}_{i=1}^{Y}} \hspace{1mm} \sum_{i=1}^{Y} \Omega_{i}^{\left(x_{i}\right)},
\end{equation}

Subject to:
\begin{equation}
\small
\sum_{i=1}^{Y} \mu_{i}^{\left(x_{i}\right)} \leq Model\hspace{1mm}Size,
\end{equation}

\begin{equation}
\small
\sum_{i=1}^{Y} \beta_{i}^{\left(x_{i}\right)} \leq BOPS \hspace{1mm}Limit,
\end{equation}

Here, $\mu_{i}^{\left(x_{i}\right)}$ denotes the size of the $i$-th layer with $x_{i}$ bit quantization and $\beta_{i}^{\left(x_{i}\right)}$ is the corresponding BOPS required for computing that layer. All the equations are adopted from \cite{yao2021hawq}. We have considered the bit precision for both weights and activations to be same so that the mathematical operations become efficient. The overall MP quantization process is summarized in figure [somethin].

\begin{figure}[htb]
\centering
\includegraphics[width=0.49\textwidth]{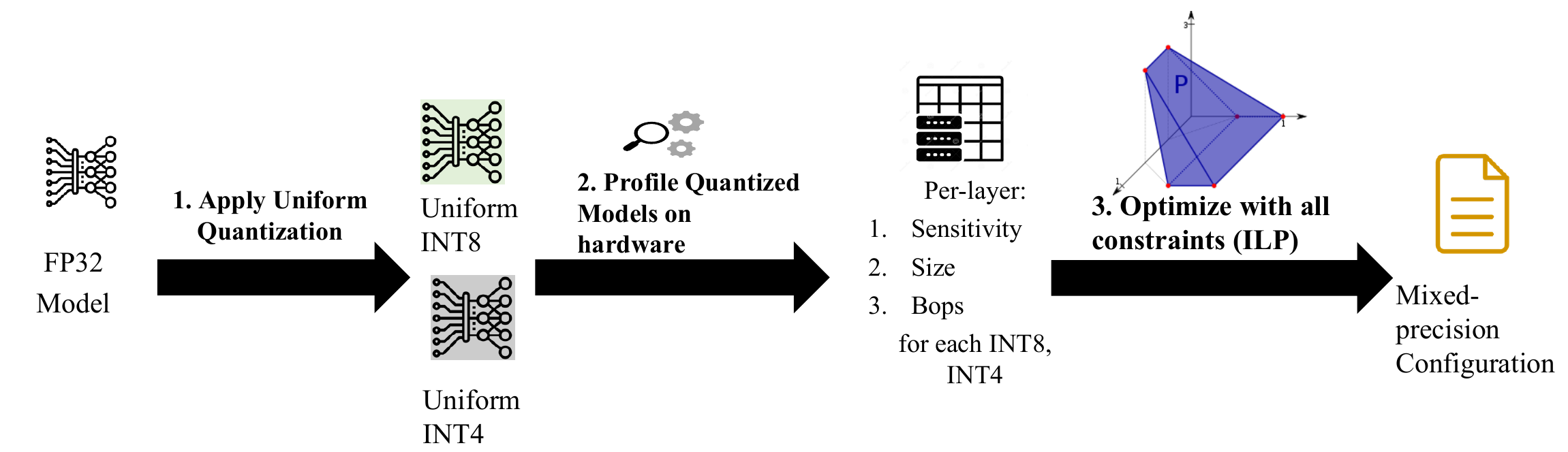}
\caption{\small Finding Mixed Precision Bit Setting Using Integer Linear Programming (ILP) }
\label{conv-dsconv}
\end{figure}

\section{\protect\sys{} Evaluation}
\label{eve}
We evaluated proposed \emph{\sys{}} with two very important real-world case studies: COVID-19 detection from multimodal audios and Battlefield Object Detection from multimodal images and audios. Both of the case studies proves to be important sectors where tinyML implementations is much required. 
\subsection{Case Study 1: Covid Detection from Multimodal Audio Recordings}

\begin{figure}[htb]
\centering
\includegraphics[width=0.4\textwidth]{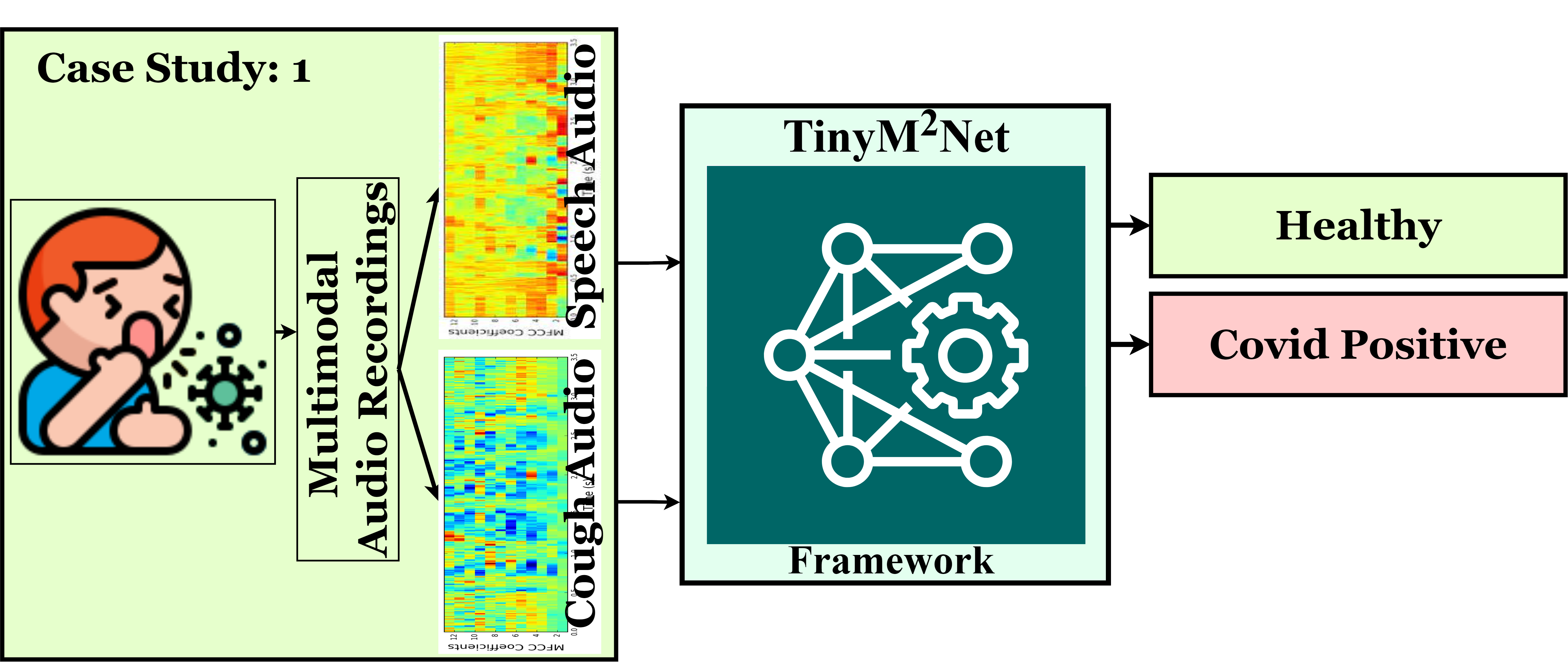}
\caption{The proposed \emph{\sys{}} framework to detect COVID-19 from two different modalities of audios, cough audios and speech audios.}
\vspace{-4ex}
\label{covid}

\end{figure}

Combining numerous data sources has always been a high priority topic, but with the advent of new AI-based learning algorithms, it has become critical to combine the complementary capabilities of distinct data sources for effective diagnosis, treatment, prognosis, and planning in a variety of medical applications. With the onset of COVID-19 pandemic, patient pre-screening from passively recorded audios has become an active area of research. Therefore, a bunch of unimodal and multimodal COVID-19 audio dataset have been presented \cite{bagad2020cough,brown2020exploring, orlandic2021coughvid,sharma2020coswara}. The ultimate goal in this research is to provide COVID-19 pre-screening mobile or tiny devices. Figure \ref{covid} shows the highlevel overview of the evaluation of \emph{\sys{}} in terms of COVID-19 detection.

\subsubsection{\textbf{Dataset Description}}
We have used the dataset from the COVID-19 cough sub-challenge (CCS) and the
COVID-19 speech sub-challenge (CSS) from the Inter Speech 2021 ComParE challenge \cite{schuller2021interspeech}. This dataset is a subset of the bigger dataset \cite{brown2020exploring} collected by University of Cambridge. In this dataset there are 929 cough audios from 397 participants and 893 speech recordings from 366 participents. Each recording included a COVID-19 test result that was self-reported by the participant. To build the two-class classification task, the original COVID-19 test results were mapped to positive (designated as `P') or negative (designated as `N') categories.
\subsubsection{\textbf{Experimental Setups, Results and Analysis}}
To create a balanced multimodal dataset we have taken 893 cough recordings from 366 participants matching their IDs mentioned in the metadata so that we have both cough and speech recordings contributing from a same person. Then we have divided the audios into 2 sec audio chunks and produces 6000 random samples out of that. Then we converted them into MFCC spectrogram and passed them to \emph{\sys{}}. \emph{\sys{}} process two different modalities of audios with its parallel CNN layers, extracts features, combines them and classify at the end as binary classification. We have used the 1st layer as traditional CNN and the later layers as DS-CNN. The detailed network architecture is mentioned in table \ref{covid_t}.

\begin{table}
\centering
\caption{\small{Details of the network architecture for multimodal COVID-19 detection}}
\label{covid_t}
\resizebox{0.48\textwidth}{!}{
\begin{tabular}{|l|l|l|} 
\hline
\multicolumn{1}{|c|}{\textbf{Layers}}  & \multicolumn{1}{c|}{\textbf{Description}}                & \multicolumn{1}{c|}{\textbf{Output}}  \\ 
\hline
Input Layer      & Cough Audio MFCC Vector             & 203$\times$20 $\times$ 1    \\ 
\hline
Input Layer                            & Speech Audio MFCC Vector                                 & 333$\times$13$\times$1                              \\ 
\hline
Conv2D           & Kernels = 16 $\times$(3$\times$3) - BN - ReLU  & 201$\times$18$\times$16    \\ 
\hline
Conv2D                                 & Kernels = 64 $\times$(3$\times$3) - BN- ReLU                           & 333$\times$13$\times$64                             \\ 
\hline
Separable Conv2D                       & Kernels = 32 $\times$(3$\times$3) - ReLU                           & 199x16x32                             \\ 
\hline
Separable Conv2D & Kernels = 32 $\times$(3$\times$3) - ReLU       & 329$\times$9$\times$32     \\ 
\hline
MaxPooling2D     & Pool size = (3$\times$3), 20\% Dropout     & 66$\times$5$\times$32      \\ 
\hline
MaxPooling2D     & Pool size = (2$\times$2), 20\% Dropout     & 164$\times$4$\times$32     \\ 
\hline
Separable Conv2D                       & Kernels = 32$\times$(3$\times$3) - ReLU                           & 64$\times$3$\times$32                           \\ 
\hline
Separable Conv2D                       & Kernels = 16$\times$(3$\times$3) - ReLU                           & 162 $\times$ 2 $\times$16                          \\ 
\hline
MaxPooling2D                           & Pool size = ($\times$3), 20\% Dropout                          & 21$\times$1$\times$32                               \\ 
\hline
MaxPooling2D                           & Pool size = (2$\times$2), 20\% Dropout                          & 81$\times$1$\times$16                               \\ 
\hline
Flatten                                & 21$\times$1$\times$32                                                  & 672                                   \\ 
\hline
Flatten                                & 81$\times$1$\times$16                                                  & 1296                                  \\ 
\hline
Dense            & Neurons = 32 - ReLU - 20\% Dropout  & 32               \\ 
\hline
Dense                                  & Neurons = 32 - ReLU - 20\% Dropout                       & 32                                    \\ 
\hline
Concatenate                            & 32 + 32                                                  & 64                                    \\ 
\hline
Dense            & Neurons = 256 - ReLU - 20\% Dropout & 256              \\ 
\hline
Dense            & Neurons = 128 - ReLU - 20\% Dropout & 128              \\ 
\hline
Dense                                  & Neurons = 2 - Softmax                                   & 2                                     \\
\hline

\end{tabular}}
\end{table}

We trained our model with categorical cross-entropy loss and Adam optimizer. We achieved 90.4\% classification accuracy with FP32 bit precision. We then quantize our model to uniform 8-bit and 4-bit precision and achieved 89.6\% and 83.6\% classification accuracy. Our MP quantization technique improves the classification accuracy to 88.4\%. The best baseline accuracies for unimodal COVID-19 detection from cough audio and speech audio were 73.9\% and 72.1\% \cite{Schuller21-TI2}. Our \emph{\sys{}} outperforms the baseline results by 14.5\%.

\subsection{Case Study 2: Battlefield Component Detection from Multimodal Images and Audios }
\begin{figure}[htb]
\centering
\includegraphics[width=0.4\textwidth]{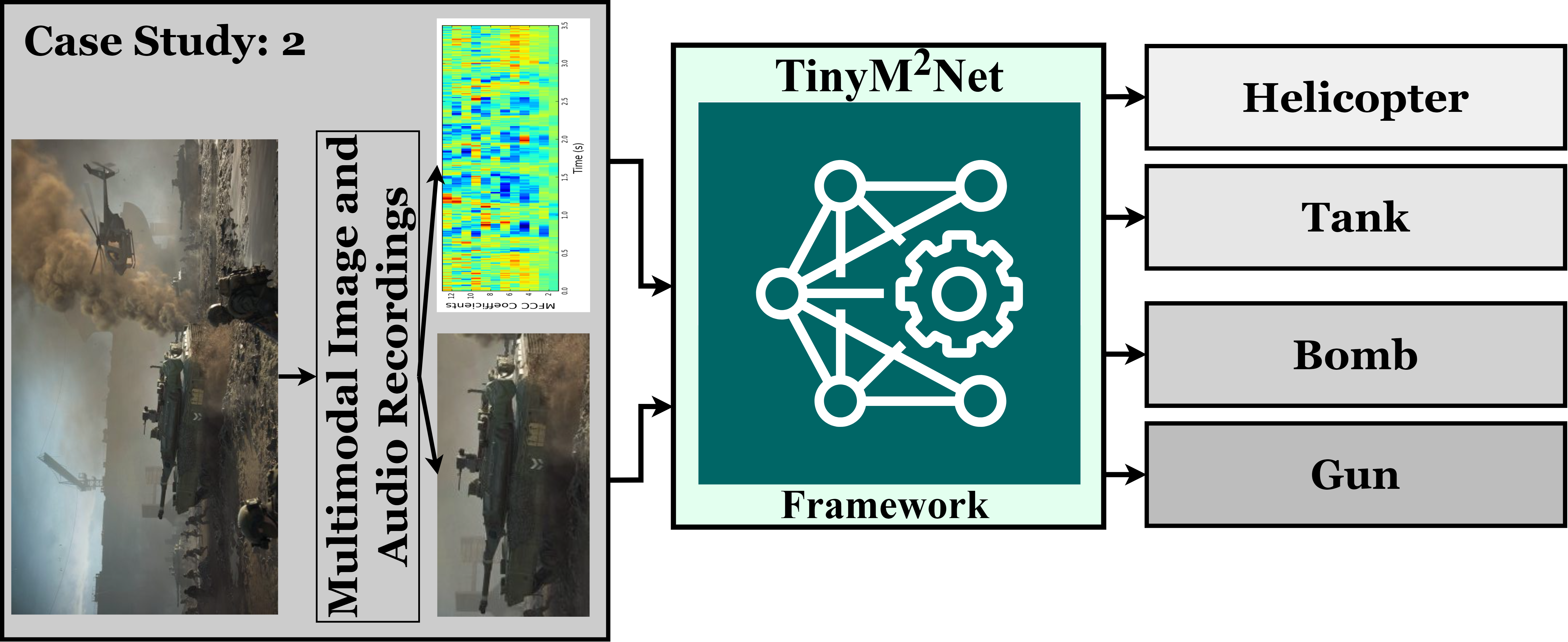}
\caption{\small The proposed \emph{\sys{}} framework to detect four different battlefield objects, Helicopter, Tank, Bomb, Gun from two different modalities, images and audios.}
\vspace{-2ex}
\label{battle}
\end{figure}

Research in the field of computer vision has always focused on the detection of specific targets in an image. Research on the identification of armored vehicles in the battlefield environment, as well as the deployment dynamics, recognition and tracking, precise strike, and so on, are critical military objectives. There are still many difficulties in detecting armored vehicles on the battlefield because of the complication of the environment \cite{xiaozhu2017object,zheng2019object}. Authors in \cite{mees2016choosing, sindagi2019mvx} proposed multimodal learning approach in object detection task. We have presented a novel multimodal learning approach in battlefield object detection based on image and audio modality. Figure \ref{battle} shows the highlevel overview of the evaluation of \emph{\sys{}} in terms of battlefield object detection. 

\subsubsection{\textbf{Dataset Description}}
As open-sourced dataset for research on battlefield environment is very limited, we have created our own dataset for this case-study. This also contributed to the novelty of our work. We have created a dataset for multiclass classification problem with 4 classes as Helicopter, Bomb, Gun, and Tank. We have selected 4 publicly available YouTube videos \cite{gshawnadams_2020,xsmasher4ya_2019, youtube_2017, youtube_2020} from where we extracted the images and corresponding audios of  Helicopter, Bomb, Gun, and Tank. Sampling rate of the image extraction was one frame per second. We have collected the images in .jpg format. We extracted in total 2745 images for all the 4 classes. On the contrary, sampling frequency was 22050Hz for 1 sec audio. Length of our audio was 1 sec. We then converted the audio signal into MFCCs. We extracted in total 2745 MFCC spectrogram for 4 classes. The total number of extracted images and corresponding audios were as follows: Helicopter-1066, Gun - 1008, Bomb - 161, Tank- 510.

\subsubsection{\textbf{Experimental Setups, Results and Analysis}}
We passed both the input and corresponding audio MFCCs to \emph{\sys{}}. \emph{\sys{}} process two different modalities with its parallel CNN layers, extracts features, combines them and classify at the end as multiclass classification. We have used the 1st layer as traditional CNN and the later layers as DS-CNN. The detailed network architecture is mentioned in table \ref{battle_t}.

We trained our model with categorical cross-entropy loss and Adam optimizer. We achieved 98.5\% classification accuracy with FP32 bit precision. We then quantize our model to uniform 8-bit and 4-bit precision and achieved 97.9\% and 88.7\% classification accuracy. Our MP quantization technique improves the classification accuracy to 97.5\% which is very comparable to both 8-bit and 32-bit quantized models. We achieved 93.6\% classification accuracy with unimodal (only image data) implementation. Our multimodal approach improved the object detection accuracy to 3.9\%.

\begin{table}
\centering
\caption{\small{Details of the network architecture for multimodal battle field object detection}}
\label{battle_t}
\resizebox{0.48\textwidth}{!}{
\begin{tabular}{|l|l|l|} 
\hline
\multicolumn{1}{|c|}{\textbf{Layers}} & \multicolumn{1}{c|}{\textbf{Description}}                                                      & \multicolumn{1}{c|}{\textbf{Output}}  \\ 
\hline
Input Layer                           & Audio MFCC Vector                                                                              & 44$\times$13$\times$1                               \\ 
\hline
Input Layer                           & Image Vector                                                                                   & 32$\times$32$\times$3                               \\ 
\hline
Conv2D                                & Kernels = 64$\times$(3 $\times$3) - BN - ReLU                                               & 44$\times$13$\times$64                              \\ 
\hline
Conv2D                                & Kernels = 64 $\times$(3$\times$3)- BN - ReLU                & 32$\times$32$\times$64                              \\ 
\hline
Separable Conv2D                      & Kernels = 32 $\times$(3$\times$3) - ReLU & 44$\times$13$\times$32                              \\ 
\hline
Separable Conv2D                      & Kernels = 32 $\times$(3$\times$3) - ReLU                  & 32$\times$32$\times$64                              \\ 
\hline
MaxPooling2D                          & Pool size = (2$\times$2), 20\% Dropout                                                                & 22$\times$6$\times$32                               \\ 
\hline
MaxPooling2D                          & Pool size = (2$\times$2), 20\% Dropout                                                                & 16$\times$16$\times$32                              \\ 
\hline
Separable Conv2D                      & Kernels = 64 $\times$(3$\times$3) - ReLU & 22 $\times$ 6 $\times$64                            \\ 
\hline
Separable Conv2D                      & Kernels = 64 $\times$(3$\times$3) - ReLU & 16 $\times$ 16 $\times$64                           \\ 
\hline
MaxPooling2D                          & Pool size = (2$\times$2), 20\% Dropout                                                                & 11$\times$3$\times$64                               \\ 
\hline
MaxPooling2D                          & Pool size = (2$\times$2), 20\% Dropout                                                                & 8$\times$8$\times$64                                \\ 
\hline
Flatten                               & 11$\times$3$\times$64                                                                                        & 2112                                  \\ 
\hline
Flatten                               & 8$\times$8$\times$64                                                                                         & 4096                                  \\ 
\hline
Dense                                 & Neurons = 64 - ReLU - 20\% Dropout                                                             & 64                                    \\ 
\hline
Dense                                 & Neurons = 64 - ReLU - 20\% Dropout                                                             & 64                                    \\ 
\hline
Concatenate                           & 64 + 64                                                                                        & 128                                    \\ 
\hline
Dense                                 & Neurons = 64 - ReLU - 20\% Dropout                                                             & 64                                    \\ 
\hline
Dense                                 & Neurons = 4 - Softmax                                                         & 4                                     \\
\hline
\end{tabular}}
\end{table}

\begin{table}[]
\caption{\small Summary of the \emph{\sys{}} Framework Evaluation Results}
\label{soft}
\scalebox{0.9}{
\begin{tabular}{|c|c|c|c|}
\hline
\textbf{\begin{tabular}[c]{@{}c@{}}Case\\ Studies\end{tabular}} & \textbf{Quantization} & \textbf{\begin{tabular}[c]{@{}c@{}}Accuracy\\ (\%)\end{tabular}} & \textbf{\begin{tabular}[c]{@{}c@{}}Model Size\\ (KB)\end{tabular}} \\ \hline
1 & Floating Points & 90.4 & 845 \\ \hline
1 & W8A8 (uniform 8) & 89.6 & 216 \\ \hline
\textbf{1} & \textbf{W4/8A4/8 (MP)} & \textbf{88.4} & \textbf{145} \\ \hline
1 & W4A4 (uniform 4) & 83.6 & 107 \\ \hline
2 & Floating Points & 98.5 & 1605 \\ \hline
2 & W8A8 (uniform 8) & 97.9 & 407 \\ \hline
\textbf{2} & \textbf{W4/8A4/8 (MP)} & \textbf{96.8} & \textbf{269} \\ \hline
2 & W4A4 (uniform 4) & 91.3 & 205 \\ \hline
\end{tabular}}
\vspace{-2ex}
\end{table}

\section{\protect\sys{} Running on Tiny Devices}
The inference stage must be implemented on resource constrained tiny devices in order to make the \emph{\sys{}} system real-time. We implemented \emph{\sys{}} on Rasberry Pi 4 which has quad-core Cortex-A72 (ARM v8) and 2GB LPDDR4 memory. Performance evaluation of the \emph{\sys{}} on resource constrained Raspberry Pi 4 was based on two metrics: inference time and power consumption during inference. The capacity of a framework to run in real time is determined by its inference time or running time. Data loading, model loading, and visual display of the final result all contribute to this inference time's length. The inference time was measured with the help of Raspbian OS's `time' function. We used a batch size of 1, which is the time it takes to process a single data point, to calculate the inference time. We also need to consider the model's power profile when deploying it in the actual world. The running power of any deep model should be well within the device's sustainable range. Power consumption is calculated by deducting idle power from the peak power indicated during inference operation and reporting the result. For reporting, we use the metric unit milliwatt (mW), and a USB power meter has been employed. The inference time and required power during inference with the most compressed models are mentioned in table \ref{hardware} for both of the case studies.
\begin{table}[]
\caption{\small Implementation of the \sys{} framework to resource constrained Raspberry Pi 4 device}
\label{hardware}
\scalebox{0.9}{
\begin{tabular}{|c|c|c|}
\hline
\textbf{Case Studies} & \textbf{\begin{tabular}[c]{@{}c@{}}Inference Time\\ (S)\end{tabular}} & \textbf{\begin{tabular}[c]{@{}c@{}}Power\\ (mW)\end{tabular}} \\ \hline
1 & 1.2 & 798 \\ \hline
2 & 1.7 & 959 \\ \hline
\end{tabular}}
\vspace{-3ex}
\end{table}

\section{Conclusion}
This paper presents \emph{\sys{}}, a flexible system algorithm co-designed multimodal learning framework which employs as much as correlated information a multimodal dataset provides in an attempt to exploit deep learning algorithms evaluating two important tinyML evaluation case-studies: to detect the signature of COVID-19 into participents' cough and speech sound and battle field object detection from multimodal images and audios. To implement into tiny hardware, extensive model compression was done in terms of networks architecture optimization and MP quantization (mixed 8-bit and 4-bit). The most compressed \emph{\sys{}} achieves 88.4\% COVID-19 detection accuracy and 96.8\% battle field object detection accuracy. Finally, we test our \emph{\sys{}} model on a Raspberry Pi 4 to see how they perform when deployed to a resource constrained tiny device. 

\section{ACKNOWLEDGMENT}

We acknowledge the support of the U.S. Army Grant No. W911NF21-20076.

\bibliographystyle{plain}
\bibliography{tinyml}

\end{document}